\documentclass{article}

     \usepackage[preprint]{template_nips/neurips_2022}

\usepackage[utf8]{inputenc} % allow utf-8 input
\usepackage[T1]{fontenc}    % use 8-bit T1 fonts
\usepackage{hyperref}       % hyperlinks
\usepackage{url}            % simple URL typesetting
\usepackage{amsfonts}       % blackboard math symbols
\usepackage{nicefrac}       % compact symbols for 1/2, etc.
\usepackage{microtype}      % microtypography
\usepackage{graphicx} % for figures
\usepackage{subcaption} % for subfigures
\usepackage{amsmath} % For math

\usepackage{amssymb} % For math symbols like \mathbb{R} etc.
\usepackage{enumitem} % For nested enumerations like 1.1., 1.2. Use: \begin{enumerate}[label*=\arabic*.]
\usepackage{latexsym} % For symbols like \ast \dagger
\usepackage[table]{xcolor} % For color in tables
\usepackage{array} % For color in tables

\newcommand\tf[1]{\textbf{#1}}
\newcolumntype{H}{>{\setbox0=\hbox\bgroup}c<{\egroup}@{}}
\newcommand\gr[1]{{\color{lightgray}{#1}}} 

\usepackage{booktabs}
\usepackage{pgfplotstable}
\usepackage{filecontents}
\usepackage{siunitx}
\usepackage{ifthen}
\newcommand\newhide[1]{}
\newboolean{show_todos}
\newboolean{show_minor_todos}
\newboolean{show_notes}
\newboolean{show_comments}
\setboolean{show_todos}{false}%{true}%
\setboolean{show_minor_todos}{false}%{true}%
\setboolean{show_notes}{true}%{false}%
\setboolean{show_comments}{false}%{true}%
\newcommand\todo[1]{\ifthenelse{\boolean{show_todos}}{\textcolor{red}{\textbf{ToDo: }#1}}{\newhide{#1}}}
\newcommand\minortodo[1]{\ifthenelse{\boolean{show_minor_todos}}{\textcolor{red}{\textbf{ToDo (Minor): }#1}}{\newhide{#1}}}
\newcommand\note[1]{\ifthenelse{\boolean{show_notes}}{\textcolor{blue}{#1}}{\newhide{#1}}}
\newcommand\comment[1]{\ifthenelse{\boolean{show_comments}}{\textcolor{blue}{\textbf{Comment:} #1}}{\newhide{#1}}}
\title{Informed Pre-Training on Prior Knowledge}

\author{%
   Laura von Rueden, Sebastian Houben, Kostadin Cvejoski, Christian Bauckhage, Nico Piatkowski \\
   Fraunhofer IAIS, University of Bonn \\
   Sankt Augustin, Germany\\
   \texttt{laura.von.rueden@iais.fraunhofer.de} \\
}

\begin{document}
\maketitle

%%%%%%%%%%%%%%%%%%%%%%%%%%%%%%%%%%%%%%%%%%%%%%%%%%%%%%%%%%%%%%%%%%%%%%%%%%%%%%%
%%%%%%%%%%%%%%%%%%%%%%%%%%%%%%%%%%%%%%%%%%%%%%%%%%%%%%%%%%%%%%%%%%%%%%%%%%%%%%%
% Paper content
%%%%%%%%%%%%%%%%%%%%%%%%%%%%%%%%%%%%%%%%%%%%%%%%%%%%%%%%%%%%%%%%%%%%%%%%%%%%%%%
%%%%%%%%%%%%%%%%%%%%%%%%%%%%%%%%%%%%%%%%%%%%%%%%%%%%%%%%%%%%%%%%%%%%%%%%%%%%%%%

\begin{abstract}
When training data is scarce, the incorporation of additional prior knowledge can assist the learning process. While it is common to initialize neural networks with weights that have been pre-trained on other large data sets, pre-training on more concise forms of knowledge has rather been overlooked. In this paper, we propose a novel informed machine learning approach and suggest to pre-train on prior knowledge. Formal knowledge representations, e.g. graphs or equations, are first transformed into a small and condensed data set of knowledge prototypes. We show that informed pre-training on such knowledge prototypes (i) speeds up the learning processes, (ii) improves generalization capabilities in the regime where not enough training data is available, and (iii) increases model robustness. Analyzing which parts of the model are affected most by the prototypes reveals that improvements come from deeper layers that typically represent high-level features. This confirms that informed pre-training can indeed transfer semantic knowledge. This is a novel effect, which shows that knowledge-based pre-training has additional and complementary strengths to existing approaches.
\end{abstract}

\section{Introduction}
The massive data requirements of neural networks have raised
important questions about how to learn from small data, how to generalize to unseen domains, and how to ensure robustness to input perturbations (\citet{wang2020generalizing, volpi2018generalizing, hendrycks2018benchmarking}).
One approach to alleviate problems due to insufficient training data is to build upon models that have been pre-trained on large related datasets.
This allows for transfer learning of general features, as shown by~\cite{yosinski2014transferable},
but task-specific concepts still need to be learned from the data at hand.
Another promising approach is to inject additional prior knowledge via informed machine learning methods, as proposed by~\cite{vonrueden21informed}.
While this ensures an alignment with semantic concepts, the integration of knowledge into learning algorithms or model architectures can be application-specific and time-consuming, which in turn raises the need for improved methods.

\begin{figure}[t]
    \centering
    \includegraphics[width=\textwidth]{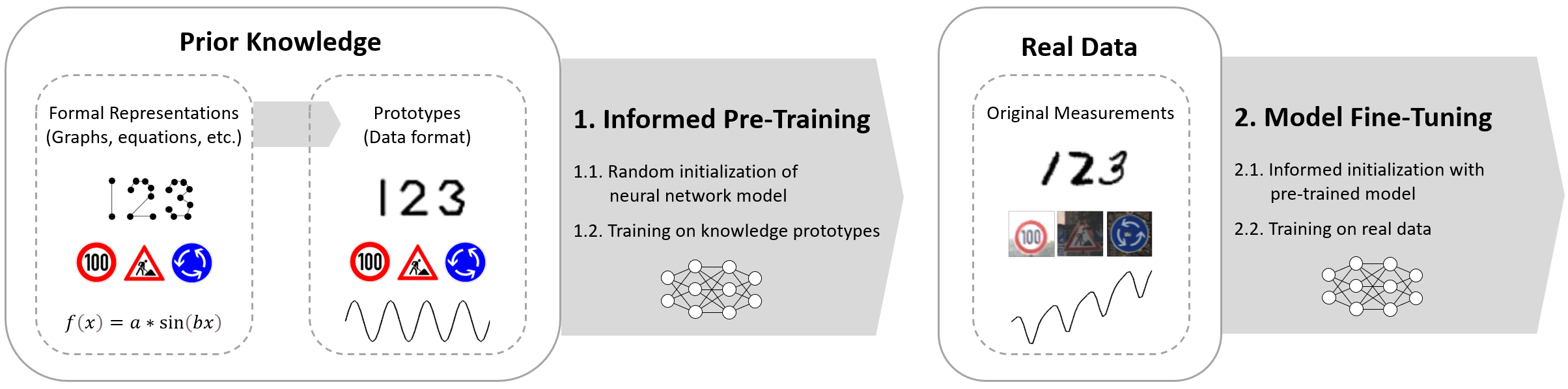}
    \caption{Approach overview.
    We propose informed pre-training on prior knowledge in order to improve neural network training, especially when real data is scarce.
    Prior knowledge is often given by formal representations, e.g. by graph structures, image templates, or scientific equations. We transform these into data format and call them "knowledge prototypes".
    This allows for informed pre-training, i.e. to train a model on prior knowledge.
    Afterwards, the pre-trained model is fine-tuned on real training data.
    The informed pre-training leads to significantly increased generalization capabilities and improved robustness.
    }
    \label{fig:approach}
\end{figure}

We propose a new informed machine learning approach and suggest pre-training on prior knowledge. This combines the advantages of both worlds: It has the powerful ability to inject well-defined and established knowledge into model training, while still being a straightforward pre-training step that can be universally applied.
Our approach is shown in Figure~\ref{fig:approach}. Given knowledge is often represented formally, e.g. by graphs or equations. We transform them into data format and then call them "knowledge prototypes". These prototypes already reflect major concepts of a target domain, so that pre-training on them leads to significant improvements.

The basic procedure is inspired by human learning:
When students learn to read digits, teachers show them a prototypical image for each category.
This lets students initially focus on the relevant information and avoids distraction by any semantically unimportant feature. 
After having internalized the main concept, the learner can simply refine it based on new observations.

Driven by the question whether 
pre-training on knowledge prototypes can improve the learning process,
we investigate
the task of image classification and consider three different datasets, namely GTSRB~(\cite{stallkamp11gtsrb}), MNIST~(\cite{lecun2010mnist}) and USPS~(\cite{uspsdataset}).
Additional experiments with the NOAA CO2 dataset~(\cite{co2data}) also show the applicability to
regression problems. 
We compare pre-training on knowledge prototypes to several baselines which go beyond default initialization and involve traditional data-based pre-training as well as other informed learning strategies.

An advantage of our method is that it can be applied to various knowledge formalization types and different domains. It is neither restricted to the types that we use in this paper (i.e. graph structures, image templates, and scientific equations) nor domain specific. Instead informed pre-training can be used for any kind of prior knowledge that can be transformed into a data format, e.g. by rendering or simulation, and might be especially beneficial for those domains where real data acquisition is hard.
For a more extensive overview of knowledge types and domains in informed machine learning we refer to the survey of~\cite{vonrueden21informed}.

The main contributions and findings of this paper can be summarized as follows:
\begin{itemize}
    \item We present pre-training on prior knowledge as a novel approach to informed machine learning. In contrast to existing pre-training approaches, our method does not rely on reusing the implicit information from large datasets but instead utilizes concise and controllable prior knowledge that is given by a small set of prototypes. To the best of our knowledge this approach has not been considered or studied before and constitutes a novel avenue for informed machine learning.
    \item Our results show that pre-training on knowledge prototypes leads to faster training convergence and can improve test accuracy for small training data by up to 11\%. Furthermore, we obtain an increase of 15\% on out-of-distribution robustness.
    \item We provide an in-depth analysis by investigating the transfer learning contribution of individual neural network layers. For traditional data-based pre-training, performance benefits arise from transferring early layers. In contrast, we observe that, for our knowledge-based pre-training, improvements stem from deeper layers which tend to represent semantic concepts. This is a novel effect which shows that pre-training on semantic features is viable and significantly different to existing approaches. We refer to this effect as ``informed transfer learning''.
    \item We compare our approach to ImageNet pre-training and find that the latter can be further improved by subsequently pre-training on knowledge prototypes. We find an additional increase of 13\% in test accuracy for small-data scenarios. This further confirms the complementary advantages of data-based and knowledge-based pre-training.
\end{itemize}

\begin{figure*}[t]%[!htb]
\centering
\begin{subfigure}[b]{0.32\textwidth}
    \includegraphics[width=\textwidth]{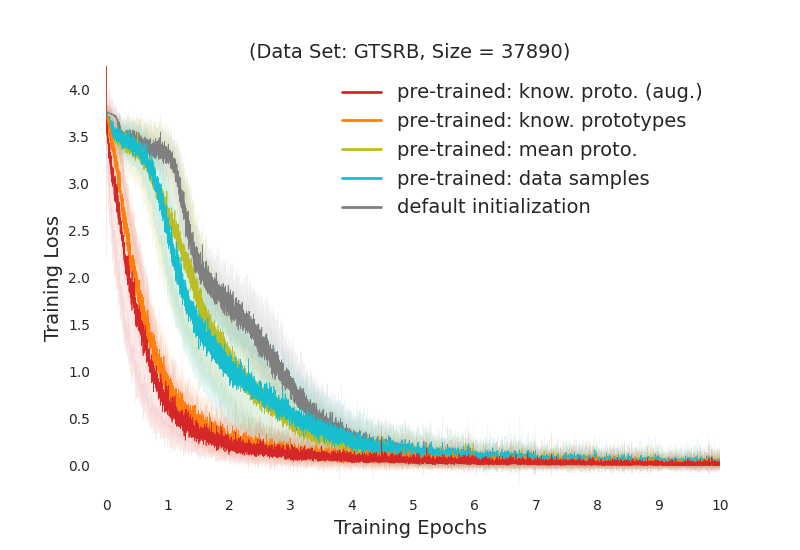}
    \caption{Training Loss}
    \label{fig:gtsrb_train_loss}
\end{subfigure}
\begin{subfigure}[b]{0.32\textwidth}
    \includegraphics[width=\textwidth]{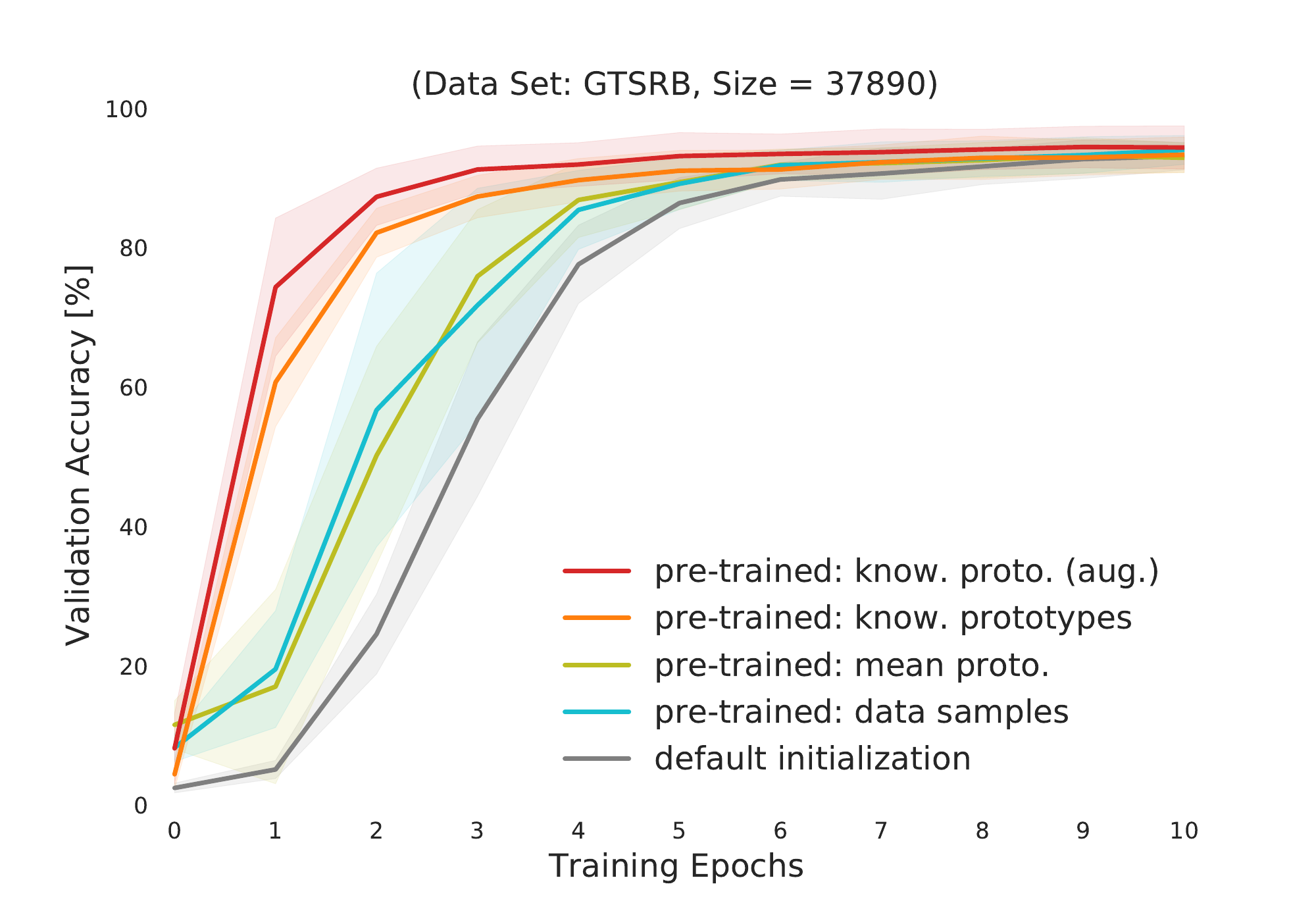}
    \caption{Validation Accuracy}
    \label{fig:gtsrb_val_acc}
\end{subfigure}
\begin{subfigure}[b]{0.32\textwidth}
    \includegraphics[width=\textwidth]{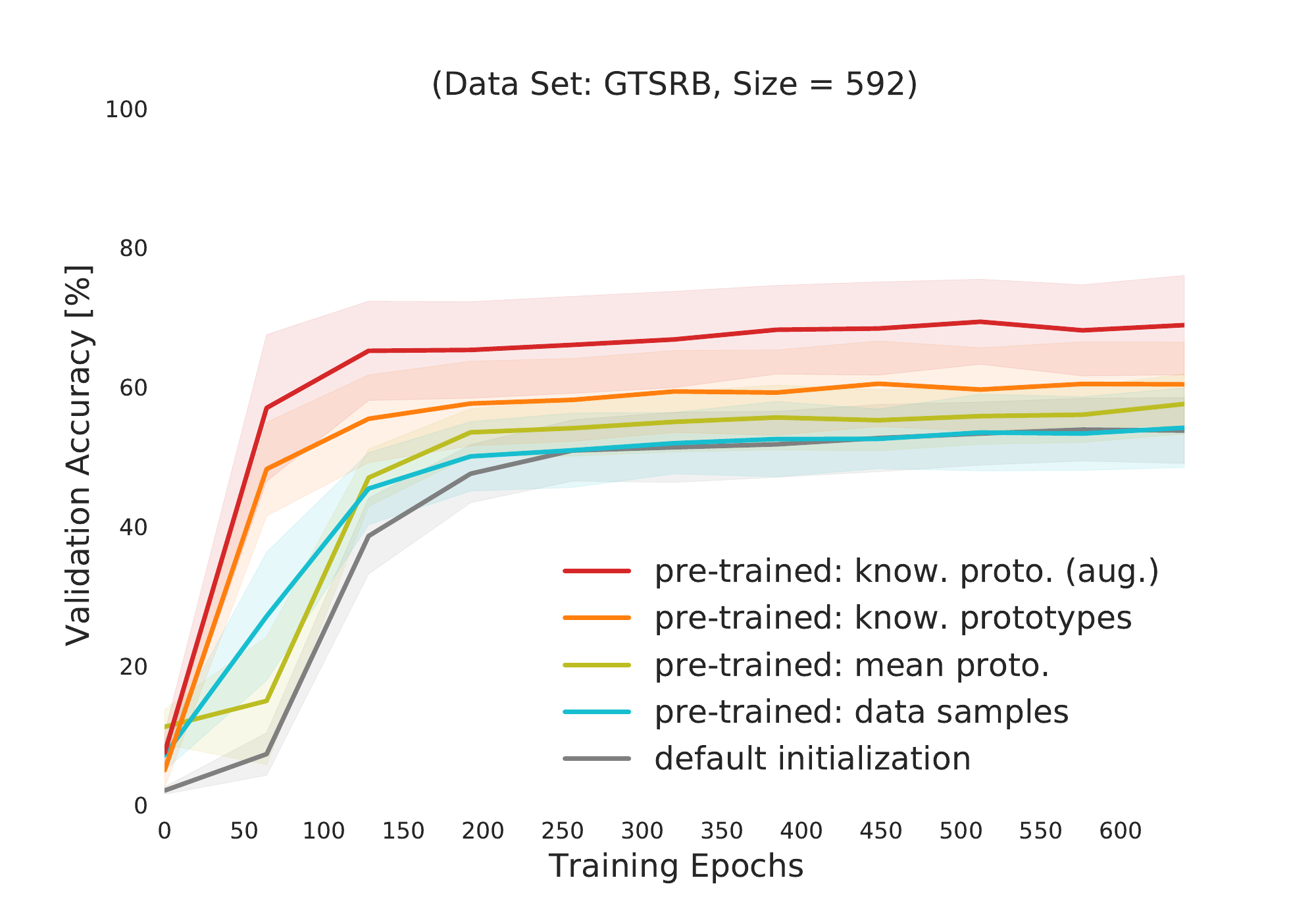}
    \caption{Val. Accuracy (Small Data)}
    \label{fig:gtsrb_val_acc_smaller_data}
\end{subfigure}
\caption{Informed pre-training on knowledge prototypes leads to a learning speed-up and improves generalization for small training data.
The training loss (\subref{fig:gtsrb_train_loss}) does not encounter local minima or saddle points but instead shows a continuous descent.
This suggests, that informed pre-training can initialize the learning algorithm in favourable regions of the loss landscape,
Moreover, for small training data the accuracy converges to a higher and better value (\subref{fig:gtsrb_val_acc_smaller_data}).
This effect can be further intensified when augmenting the knowledge prototypes.
}
\label{fig:gtsrb_acc}
\end{figure*}

\section{Related Work}
Our paper is located at the intersection of informed machine learning, pre-training, and transfer learning and also relates to the use of prototypes in artificial intelligence.

\paragraph{Informed Machine Learning.}

Informed learning describes the idea to improve data-based learning systems through the integration of additional prior knowledge~(\cite{vonrueden21informed}).
The utilization of prior information for regularization has already been discussed many years ago in the context of statistical learning theory~(\cite{vapnik1995nature, niyogi1998incorporating}).
Recently, knowledge injection into neural networks became a popular approach to alleviate the problem of small training data or to ensure knowledge conformity.
As outlined by the taxonomy of informed machine learning, given knowledge representations can be integrated into the machine learning pipeline using different approaches~(\cite{vonrueden21informed}).
For example, logic rules and physical equations can be integrated into the loss function as constraints~(\cite{diligenti2017semantic, stewart2017label}), or they can be incorporated into the model architecture as inductive biases~(\cite{battaglia2018relational, karniadakis2021physics}).
Recently it has also been shown that logic rules describing object shapes can be used to improve adversarial robustness~(\cite{gurel2021knowledge}).

\paragraph{Pre-Training \& Transfer Learning.}

The weights of a neural network are usually initialized randomly~(\cite{glorot2010understanding, he2015delving}) but can also be initialized by reusing the parameters from a pre-trained model.
It can then be fine-tuned on given training data for the target task.
\citet{erhan2009difficulty} studied the advantages of unsupervised pre-training and found that it leads to better performing classifiers and is beneficial with small training datasets. They concluded that this is not only an improved optimization procedure but also leads to better generalization.
In the last years, supervised pre-training on the ImageNet dataset~(\cite{deng2009imagenet}) became a common transfer learning approach, especially for computer vision tasks~(\cite{yosinski2014transferable,huh2016makes, neyshabur2020being}).
This approach is also used in the context of few-shot learning, i.e. generalizing from only a few examples~(\cite{wang2020generalizing}).
\citet{he2019rethinking} put the advantages of pre-training with ImageNet into question and claimed that it does not necessarily result in better test performance but only in a speed-up of the learning process. Nevertheless, \citet{hendrycks2019using} showed that pre-training can improve robustness. 

Several of the above mentioned works also studied layer transferability~(\cite{erhan2009difficulty, yosinski2014transferable, neyshabur2020being}). They found that the largest benefits in performance stem from pre-training the early layers. These often represent general features and low-level statistics of the data.

\paragraph{Prototypes.}

In the context of machine learning, prototypes are representatives of a data distribution.

Such prototypes can be available from prior knowledge.
This is quite natural in the computer vision domain, where prototypical images or structural prototypes are traditionally used as templates for object classification.
For example, \citet{hastie1994handwritten} used deformable prototypes for handwritten digit recognition, or \citet{larsson2011using} used spatial prototypes for traffic sign recognition.
Prototypical images are also used in modern approaches, especially when dealing with small datasets.
They can, for example, be the basis for creating synthetic training data~(\cite{spata2019generation}, or be employed with autoencoders for one-shot learning~(\cite{kim2019variational}.

When prototypes are not yet available, they can also be learned from the dataset.
Prototypical data elements can be automatically identified and then used for analysis of model explainability (\cite{kim2016examples, li2018deep, chen2019looks}.
It is also possible to identify such prototypes in the latent space and use them implicitly within so called prototypical networks~(\cite{snell2017prototypical}.

Moreover, there are also parallels between our work and curriculum or continual learning (\cite{bengio2009curriculum, mobahi2015theoretical}). The main difference is that curriculum learning usually uses a subset of training examples, not synthetic prototypes.

\begin{figure*}[t]%[!htb]
\centering
\begin{subfigure}[b]{0.49\textwidth}
    \includegraphics[width=\textwidth]{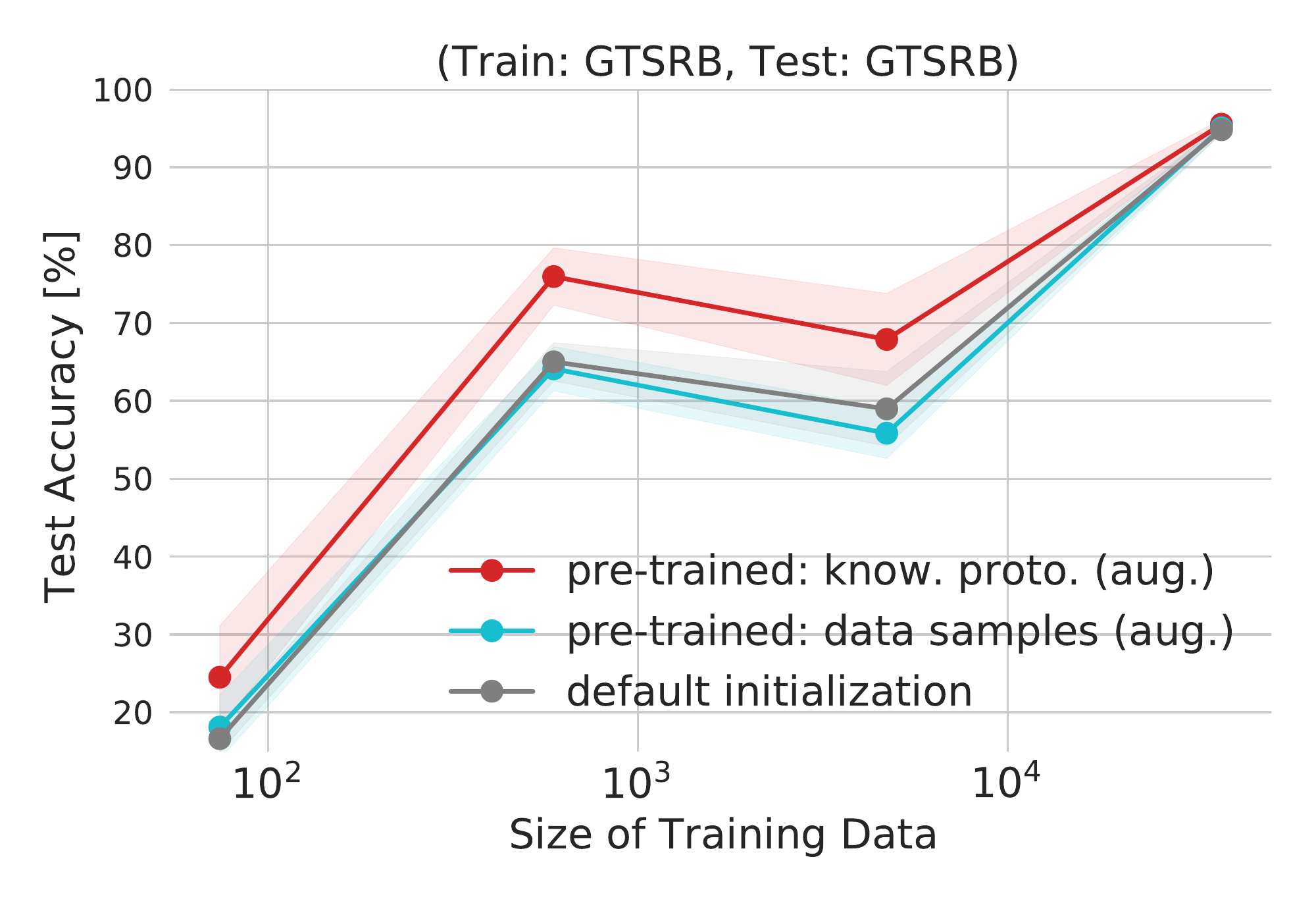}
    \caption{GTSRB}
    \label{fig:gtsrb_acc_vs_size}
\end{subfigure}
\begin{subfigure}[b]{0.49\textwidth}
    \includegraphics[width=\textwidth]{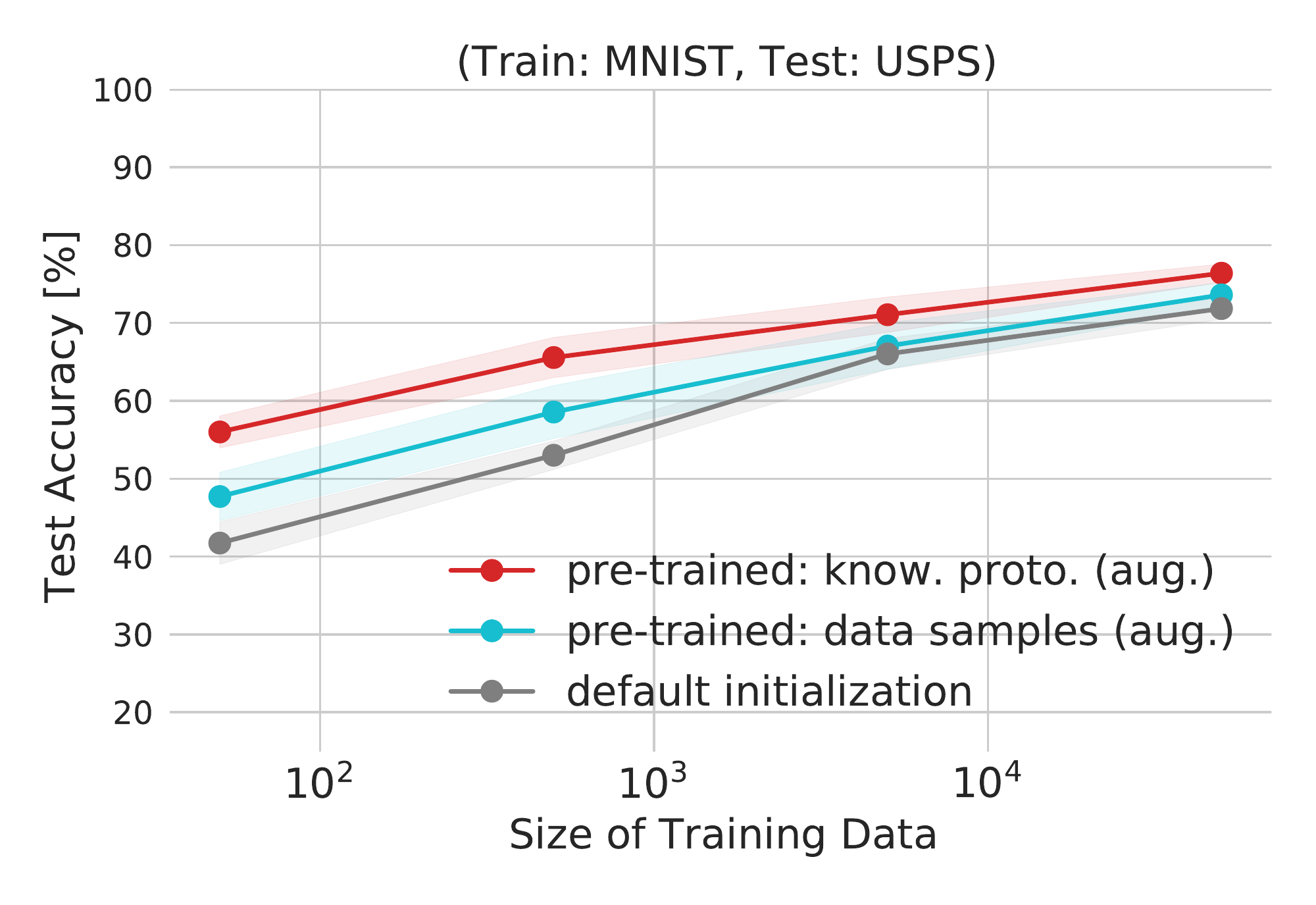}
    \caption{USPS}
    \label{fig:usps_acc_vs_size}
\end{subfigure}
\caption{
Pre-training on knowledge prototypes improves generalization for small training data, as well as robustness for all data set sizes.
The figures show the test accuracies for different training data sizes, which account for 0.1\%,  1\%,  10\%, and 100\% of the full data set.
Informed pre-training consistently improves the test accuracy for smaller data subsets.
Figure~(\subref{fig:usps_acc_vs_size}) shows out-of-distribution robustness.
Here, informed pre-training improves even the results on the full data set.
% large error bars for small data sets are due to shuffling the data subset for each repetition
}
\label{fig:acc_vs_size}
\end{figure*}
%%%%%%%%%%%%%%%%%%%%%%%%%%%%%%%%%%%%%%%%%%%%%%%%%%%%%%%%%%%%%%%%%%%%%%%%%%%%%%%%%%%%%%%%%%%%%%%%%%%%%%%%%%%%%%%%%%%%%%
%%%%%%%% Test on GTSRB + MNIST + USPS %%%%%%%%%%%%%%%%%%%%%%%%%%%%%%%%%%%%%%%%%%%%%%%%%%%%%%%%%%%%%%%%%%%%%%%%%%%%%%%%%%%%%%%%%%%%%%%
%%%%%%%%%%%%%%%%%%%%%%%%%%%%%%%%%%%%%%%%%%%%%%%%%%%%%%%%%%%%%%%%%%%%%%%%%%%%%%%%%%%%%%%%%%%%%%%%%%%%%%%%%%%%%%%%%%%%%%

%\begin{table*}[!htb]
\begin{table}[t]%[!htb]
\centering
\footnotesize
%\scriptsize
\begin{filecontents*}{tables/results_tex_table_alldata_ep_10_acc_vs_trainset_size_ONLY_PRTR_one_col_no_highlighting.csv}
test data,pre-training,0.1,1,10,100
GTSRB,(default init.),16.61 \pm 1.74,65.01 \pm 2.46,58.97 \pm 4.79,94.82 \pm 0.48
,know. prototypes,19.87 \pm 3.21,69.81 \pm 3.89,62.20 \pm 5.07,95.07 \pm 0.73
,know. proto. (aug.),24.50 \pm 6.61,75.96 \pm 3.69,67.89 \pm 5.91,95.53 \pm 0.62
,data samples,18.18 \pm 2.96,65.38 \pm 2.89,61.60 \pm 3.13,94.89 \pm 0.57
,data samples (aug.),17.69 \pm 4.15,64.16 \pm 3.01,55.39 \pm 3.07,95.07 \pm 0.72
,mean prototypes,17.45 \pm 3.62,65.69 \pm 4.57,58.47 \pm 6.54,94.82 \pm 0.79
MNIST,(default init.),73.00 \pm 3.04,89.55 \pm 1.12,97.25 \pm 0.16,98.53 \pm 0.11
,know. prototypes,72.51 \pm 2.70,89.90 \pm 1.12,97.26 \pm 0.15,98.46 \pm 0.18
,know. proto. (aug.),77.57 \pm 4.26,91.77 \pm 0.81,97.16 \pm 0.23,98.69 \pm 0.13
,data samples,73.38 \pm 3.11,89.93 \pm 0.62,97.16 \pm 0.15,98.46 \pm 0.18
,data samples (aug.),75.43 \pm 4.09,90.96 \pm 0.77,97.08 \pm 0.26,98.54 \pm 0.14
,mean prototypes,69.81 \pm 3.47,89.53 \pm 1.19,97.08 \pm 0.26,98.42 \pm 0.22
USPS,(default init.),41.73 \pm 2.73,53.01 \pm 1.83,66.03 \pm 1.99,71.85 \pm 1.38
,know. prototypes,42.11 \pm 4.82,55.06 \pm 2.31,66.93 \pm 1.81,72.68 \pm 1.33
,know. proto. (aug.),56.00 \pm 2.05,65.56 \pm 2.58,71.06 \pm 2.27,76.39 \pm 1.18
,data samples,41.75 \pm 3.50,52.45 \pm 2.83,64.98 \pm 1.58,72.54 \pm 1.83
,data samples (aug.),47.71 \pm 3.09,58.56 \pm 3.40,67.04 \pm 2.99,73.61 \pm 1.56
,mean prototypes,39.70 \pm 3.24,52.10 \pm 2.54,64.76 \pm 2.66,71.78 \pm 1.55
\end{filecontents*}
\pgfplotstabletypeset[
    col sep=comma,
    ignore chars={\$},% siunitx does not like $\pm$, but \pm is fine
    string type,
    % multicolumn header
    every head row/.style={before row={\toprule &  & \multicolumn{4}{c}{\tf{Test Accuracies [\%], for different Train Data Sizes}}\\}, after row=\midrule,},
    % Define header 
    multicolumn names={l},% do not apply S format to header
    display columns/0/.style={column type = {l},column name={\tf{Test Data}}},
    display columns/1/.style={column type = {l},column name={\tf{Pre-Training}}},
    display columns/2/.style={column type = {S},column name={\hspace{0.6em}\tf{$\approx$ 0.1\%}}},
    display columns/3/.style={column type = {S},column name={\hspace{0.6em}\tf{$\approx$ 1\%}}},
    display columns/4/.style={column type = {S},column name={\hspace{0.6em}\tf{$\approx$ 10\%}}},
    display columns/5/.style={column type = {S},column name={\hspace{0.6em}\tf{100\%}}},
    % rules
    every row no 5/.style={after row=\midrule},
    every row no 11/.style={after row=\midrule},
    every last row/.style={after row=\bottomrule},
    % grey fonts
    every row no 0 column no 1/.style={postproc cell content/.style={@cell content=\gr{##1}}},
    every row no 6 column no 1/.style={postproc cell content/.style={@cell content=\gr{##1}}},
    every row no 12 column no 1/.style={postproc cell content/.style={@cell content=\gr{##1}}},
    % textbf for know. proto. labels
    every row no 1 column no 1/.style={postproc cell content/.style={@cell content=\tf{##1}}},
    every row no 2 column no 1/.style={postproc cell content/.style={@cell content=\tf{##1}}},
    every row no 7 column no 1/.style={postproc cell content/.style={@cell content=\tf{##1}}},
    every row no 8 column no 1/.style={postproc cell content/.style={@cell content=\tf{##1}}},
    every row no 13 column no 1/.style={postproc cell content/.style={@cell content=\tf{##1}}},
    every row no 14 column no 1/.style={postproc cell content/.style={@cell content=\tf{##1}}}
%]{tables/results_tex_table_alldata_ep_10_acc_vs_trainset_size_ONLY_PRTR.csv}
]{tables/results_tex_table_alldata_ep_10_acc_vs_trainset_size_ONLY_PRTR_one_col_no_highlighting.csv}
\caption{
Test accuracies in \% (higher is better) for different training data sizes.
We report the mean values and their standard deviations from 10 repetitions.
The bold rows highlight the results based on pre-training on knowledge prototypes.
The first presents pre-training on only 1 prototype per class, and the second (aug.) presents the results for augmentation with 100 prototypes per class.
The same augmentation procedure is applied to the reference method of pre-training on a few random data samples.
%Underlined results highlight the best results for each test case and training data size.
The first case was trained and tested on GTSRB, the second on MNIST (distinct train / test subsets). The third case was trained on MNIST, but tested on USPS and highlights the improvements in out-of-distribution robustness.
}
\label{tab:results_tab}
\end{table}

\section{Informed Pre-Training on Prior Knowledge}
We investigate the effect of pre-training on prototypes derived from prior knowledge.
In the following, we briefly describe the practical approach.
A mathematical formalization that further motivates the approach can be found in the Appendix.

From a practical point of view our approach consists of the following two main phases, as illustrated in Figure~\ref{fig:approach}:

\begin{enumerate}
    \item Model pre-training
    \begin{enumerate}[label*=\arabic*.]
        \item Initialization of neural network model
        \item Training on knowledge prototypes
    \end{enumerate}
    \item Model fine-tuning
    \begin{enumerate}[label*=\arabic*.]
        \item Informed initialization with pre-trained model
        \item Training on real data
    \end{enumerate}
\end{enumerate}

In the first phase, the model is trained only on the prototypes.
Taking into account that the number of model parameters is much bigger then the number of samples, we consciously let the model memorize the prototypes and run the learning algorithm until the training loss approaches zero.
In this situation this is favourable because the prototypes contain no noise and we want to fully exploit the prior knowledge.
We further support this design choice with the recent findings about the double-descent risk curve, which describes the effect that in an over-parameterized regime a decreasing test error can be observed~(\cite{belkin2019reconciling, nakkiran20deep}).

In the second phase, the model is initialized with the pre-trained model and then fine-tuned to the original training data.
For both phases, we use the same model architecture and the same hyperparameters.

\section{Experimental Setup}
% Datasets
We demonstrate our approach with experiments on the task of image classification.
We employ three different datasets, which are 
the German Traffic Sign Recognition Benchmark (GTSRB)~(\cite{stallkamp11gtsrb}),
the widespread handwritten digit database from MNIST~(\cite{lecun2010mnist}),
and the related but different digit dataset USPS~(\cite{uspsdataset}) in order to test for out-of-distribution robustness.
We have also done experiments with the NOAA dataset of atmospheric carbon dioxide, CO2~(\cite{co2data}), which can be found in the Appendix. They show that our approach can also be applied to other tasks.

% Models
Our goal was to show the benefits of our method with well-known benchmark datasets and standard architectures and hyperparameters.
Therefore, in this regard we relied on pre-given standards.
For GTSRB we used the AlexNet~(\cite{krizhevsky2012imagenet}) and for MNIST the LeNet-5 convolutional neural network architecture~(\cite{lecun1998gradient}).
We have also tested our method on several additional architectures, to confirm that the observed effects generalize to other setups. 
These additional results \minortodo{Add to appendix}, as well as more details on the datasets, as well as the used model architectures, hyperparameters, learning algorithms, and computing infrastructure are given in Appendix.
%in Section~\ref{sec:exp_setup_appendix}.

% Repetitions
We repeat every experiment run for 10 times, and for every run we redo both the pre-training and the fine-tuning.
The repetitions differ in the random initialization of the pre-training phase, as well as in the split into training and validation data for the fine-tuning phase.
% Epochs
We run the fine-tuning for a fixed budget of 10 training epochs and apply no early stopping or other regularization.
Although the 10 epochs are enough to reach a sufficient convergence, we also run extended experiments for 100 epochs for which results can be found in Appendix.% in Table~\ref{tab:results_tab_100_epochs}.

% Dataset sizes
We evaluate the results for four different training data sizes: 100\%, 10\%, 1\%, and 0.1\%.
For MNIST this means: 50,000 / 5,000 / 500 / 50 data items. For GTSRB we create the shares through dividing by 8: 37,890 / 4710 / 592 / 74.
The exact numbers result from maintaining the organization into image tracks (30 images each) for the larger two, and permitting the use of individual images for the smaller two data subsets. The number of training epochs is increased anti-proportionally to keep the number of totally seen data elements per training comparable for all sizes.
The smallest data subsets can be considered to be in the few shot data regime, because it contains only a handful of elements per label.

\subsection{Pre-Training on Prototypes}

We run the pre-training on a small set of only one prototype per class.
After fine-tuning on the original training data we evaluate the accuracy on the test data.
We compare the results based on pre-training on knowledge prototypes to default initialization but also to pre-training on the same number of data samples.
This allows us to differentiate the benefits of informed transfer learning from the pure advances from the learning algorithm.
We also compare pre-training on alternative prototypes, which we compute as the mean image for every class.
Finally, we also investigate geometric augmentation of the prototypes from 1 up to  100 prototypes per class.

For GTSRB, we utilized publicly available templates of traffic sign symbols for the knowledge prototypes. In fact, we simply recycled the symbols from the official GTSRB result analysis application. A subset of 10 from the original 43 classes can be found in the Appendix.% in Figure~\ref{fig:gtsrb_prototypes}.
For MNIST, we employ the deformable graph prototypes from~\citet{hastie1994handwritten}. For this we simply made a screenshot and increased the line width so that edges and nodes are smooth and transform them into images. 
These are depicted in the Appendix.% in Figure~\ref{fig:mnist_prototypes}.

\section{Results}
\label{sec:results}

Our experiments show that informed pre-training on knowledge prototypes leads to a learning speed-up, improved generalization for small training data, and to improved robustness.
We observe that these benefits result from transfer learning of late layers.
Finally, we compare informed pre-training to other informed learning strategies.
In the following we present the results on each of these findings.

\subsection{Learning Speed-Up}
\label{sec:51_speed_up}
Figure \ref{fig:gtsrb_acc} depicts the learning curves for our experiments on the GTSRB dataset.
Figure \ref{fig:gtsrb_train_loss} shows the training loss. When using the default initialization, it exhibits a bumpy pattern and only converges irregularly.
This suggests that several distinct local minima are traversed before settling on the final parameters.
In contrast, after pre-training on knowledge prototypes, the loss shows a direct and smooth convergence.
This indicates that the model is initialized in a more favourable region of the loss landscape.
We further compare our approach to pre-training on data samples. Although it leads to a slight speed-up, pre-training on knowledge prototypes is significantly faster. This shows that the benefits are not only due to the higher number of training iterations but indeed to the use of knowledge prototypes.
The learning speed-up is also reflected in a better validation accuracy after fewer iterations, as shown in Figure~\ref{fig:gtsrb_val_acc}.
For small training data, the accuracy reaches a higher value, as depicted in Figure~\ref{fig:gtsrb_val_acc_smaller_data} and further discussed in Section~\ref{sec:52_generalization}.
Learning curves for MNIST can be found in the Appendix%. in Figure~\ref{fig:mnist_acc}.
All in all, these accuracy improvements, i.e., a improved start, a steeper slope and a higher asymptote, agree with those that can generally be expected from successful transfer learning, as summarized by~\cite{torrey2010transfer}.

The initial loss in the very first fine-tuning iteration after pre-training can be larger than from random initialization, but it immediately recovers and reaches a better minimum.
We interpret this as follows: Pre-training can lead to an initialization in a region of the loss function that is unlikely to reach by the learning algorithm, but which is already close to the optimum.
This effect is clearly visible in the respective Figure
%in Figure \ref{fig:gtsrb_val_loss_moved}
in the Appendix.
We argue that pre-training on knowledge prototypes reflects the overall structure of the main learning task.
In particular, pre-training is an easier to solve problem and that the model that minimizes the training loss on the knowledge prototypes is close to the one that minimizes the training loss on the target data.
This interpretation can be likened to continuation methods, where a non-convex optimization problem is transformed into a convex optimization problem and then gradually converted back while following the path of the minimizer~(\cite{bengio2009curriculum, mobahi2015theoretical}).

\subsection{Improved Generalization and Robustness}
\label{sec:52_generalization}
One motivation for the integration of prior knowledge into machine learning is to alleviate the problem of little training data.
We thus run experiments for different training data sizes in the model fine-tuning step.
Figure~\ref{fig:acc_vs_size} visualizes the resulting test accuracies and
Table~\ref{tab:results_tab} shows the complete results.
For GTSRB, informed pre-training improves the final test accuracy by around 8-11\% compared to default initialization for the small training data sizes (0.1-10\% of the original data, i.e.~, from below 100 to a few thousand images).
For MNIST, the improvements for the smallest data set are around 4-5\%. However, here, the default test accuracy is already on a quite high level although training is performed with only 50 data elements. Obviously, the recognition of greyscale digits is a much simpler problem then the classification of traffic signs.
The test accuracies after training on full data sets are slightly better for both GTSRB and MNIST but lie within one standard deviation of the default.
This is in line with the findings from~\cite{he2019rethinking} that for large datasets, training from random initialization can converge to the pre-trained solution.

As a side note, please note that for GTSRB the zig-zag pattern is due to the data set organization itself. The 0.1\% and 1\% data subsets consist of individual images, whereas the 10\% and 100\% data consists of image scene tracks with 30 elements. Therefore the 1\% data subset contains more different scenes and can lead to a better test accuracy.

Figure~\ref{fig:usps_acc_vs_size} shows the out-of-distribution robustness for models trained on MNIST with respect to their accuracy on the USPS dataset~(\cite{uspsdataset}). 
This dataset consists of handwritten digit images as well, but they stem from a different underlying data distribution.
Potential domain shifts between training and target data are a challenge in machine learning and it is desirable that a trained model is robust to out-of-distribution data.
In our experiments, we observe that pre-training on knowledge prototypes improves the accuracy for the smallest data set by nearly 15\%.
What is even more remarkable is that it also improves the model that was trained on the full data set by about 5\%.
This substantiates that our proposed approach can significantly improve robustness. 

\begin{figure}[t]%[htb]%[h!tb]
\centering
\begin{subfigure}[b]{0.32\columnwidth}
    \includegraphics[width=\textwidth]{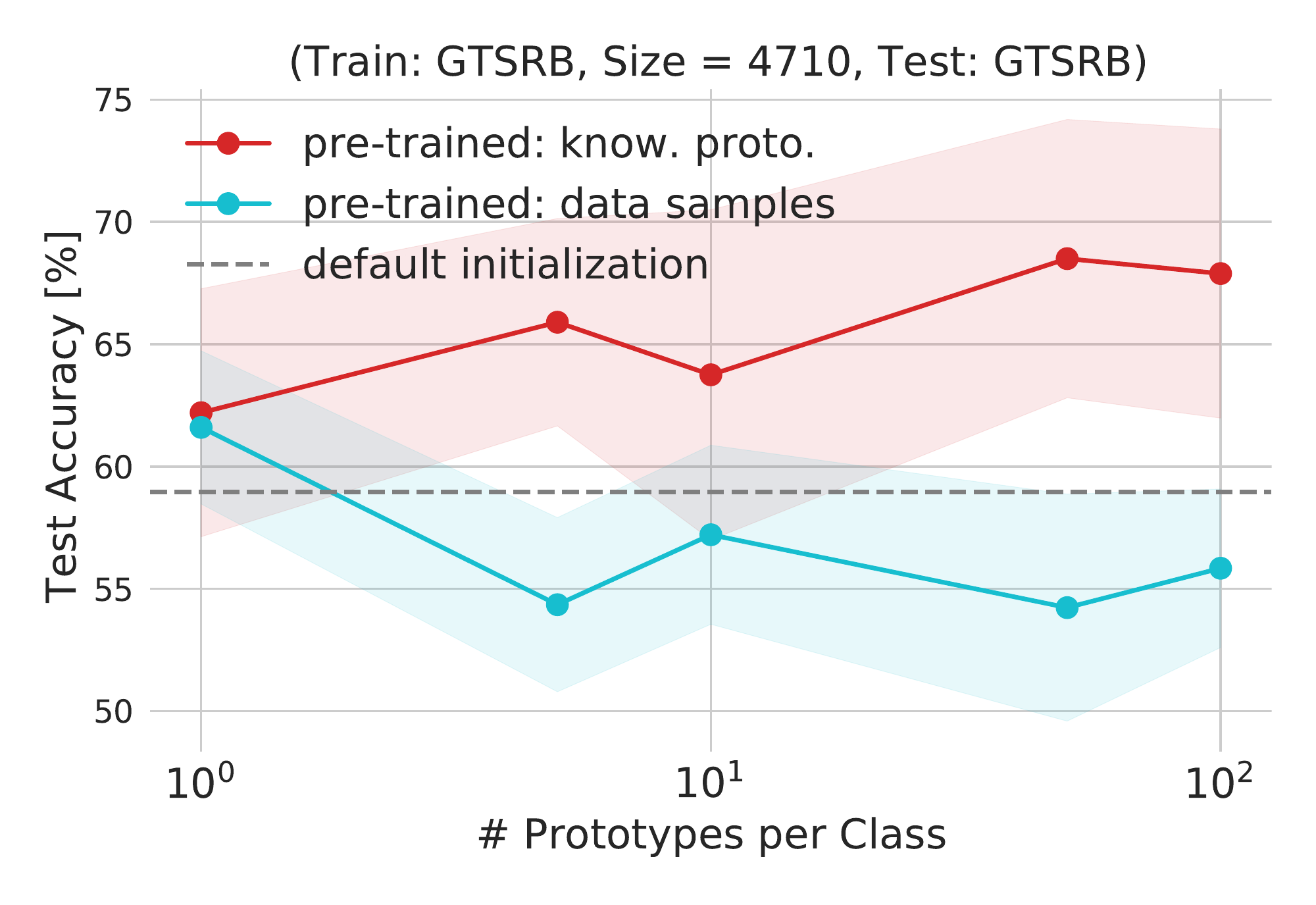}
    \caption{GTSRB}
    \label{fig:gtsrb_acc_vs_aug}
\end{subfigure}
\begin{subfigure}[b]{0.32\columnwidth}
    \includegraphics[width=\textwidth]{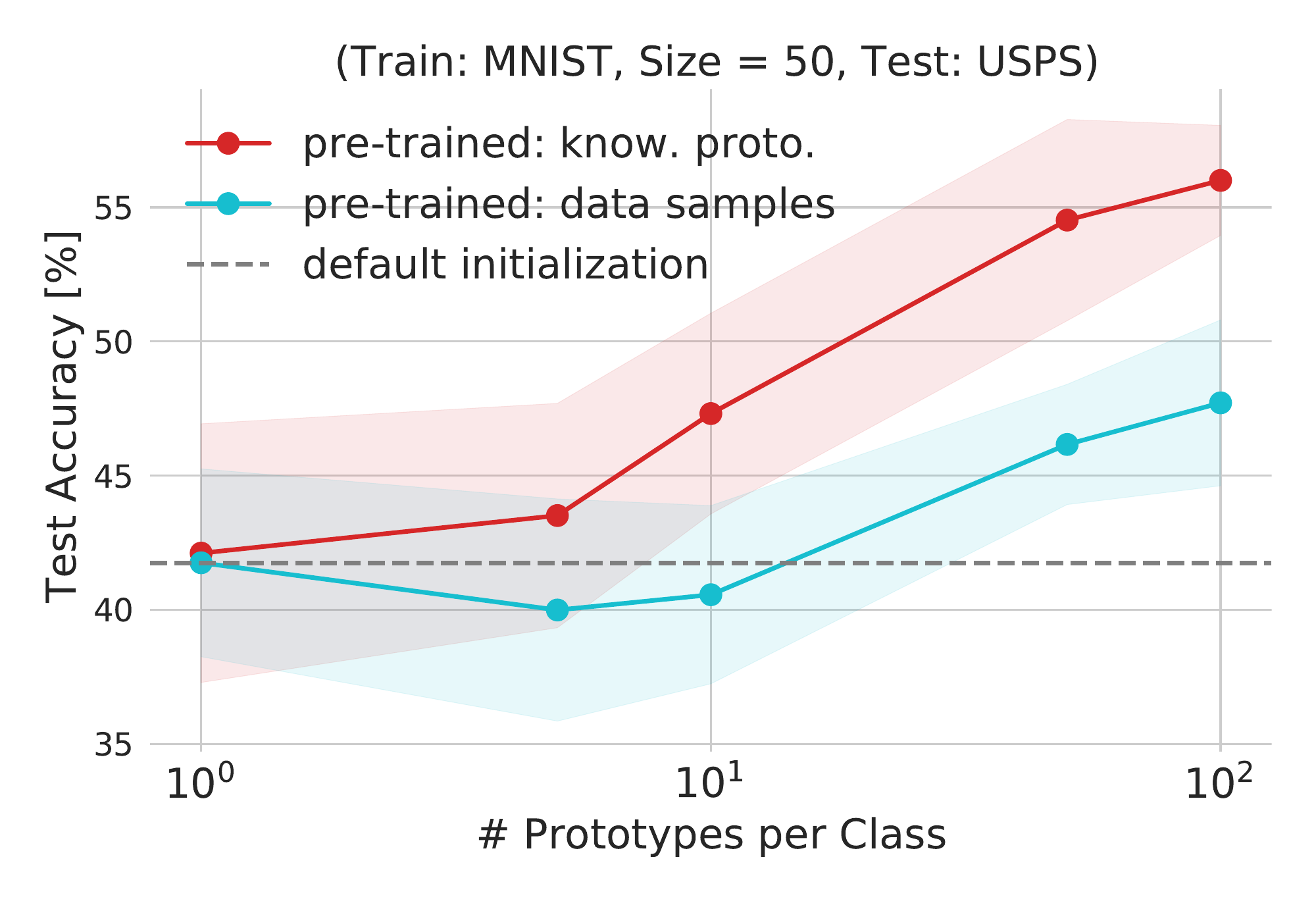}
    \caption{USPS}
    \label{fig:mnist_acc_vs_aug}
\end{subfigure}
\caption{
Test accuracy after pre-training with respect to number of prototypes per class. We started with 1 prototypes per class, then applied augmentations up until 100 prototypes per class.
}
\label{fig:acc_vs_aug}
\end{figure}

We investigated augmentation of the knowledge prototypes. Although even pre-training on only one prototype per class leads to improvements, it can be significantly increased through geometric augmentations.
MNIST prototypes were augmented with random 2D perspective transformations (e.g. rescaling, rotation, shearing), GTSRB prototypes with plausible random affine transformations: Rescaling, small translations, rotation up to 5 degree.
Figure~\ref{fig:acc_vs_aug} shows that augmenting knowledge prototypes is highly successful, whereas augmenting data samples can even lead to deterioration. 
More results on the augmentation effect can be found in the Appendix%. in Figure~\ref{fig:acc_vs_aug_all}.

\subsection{Transfer Learning of Semantic Layers}
\label{sec:53_transfer}
\begin{figure*}[t]%[h!tb]
\centering
\begin{subfigure}[t]{0.32\textwidth}
    \centering
    \includegraphics[width=\textwidth]{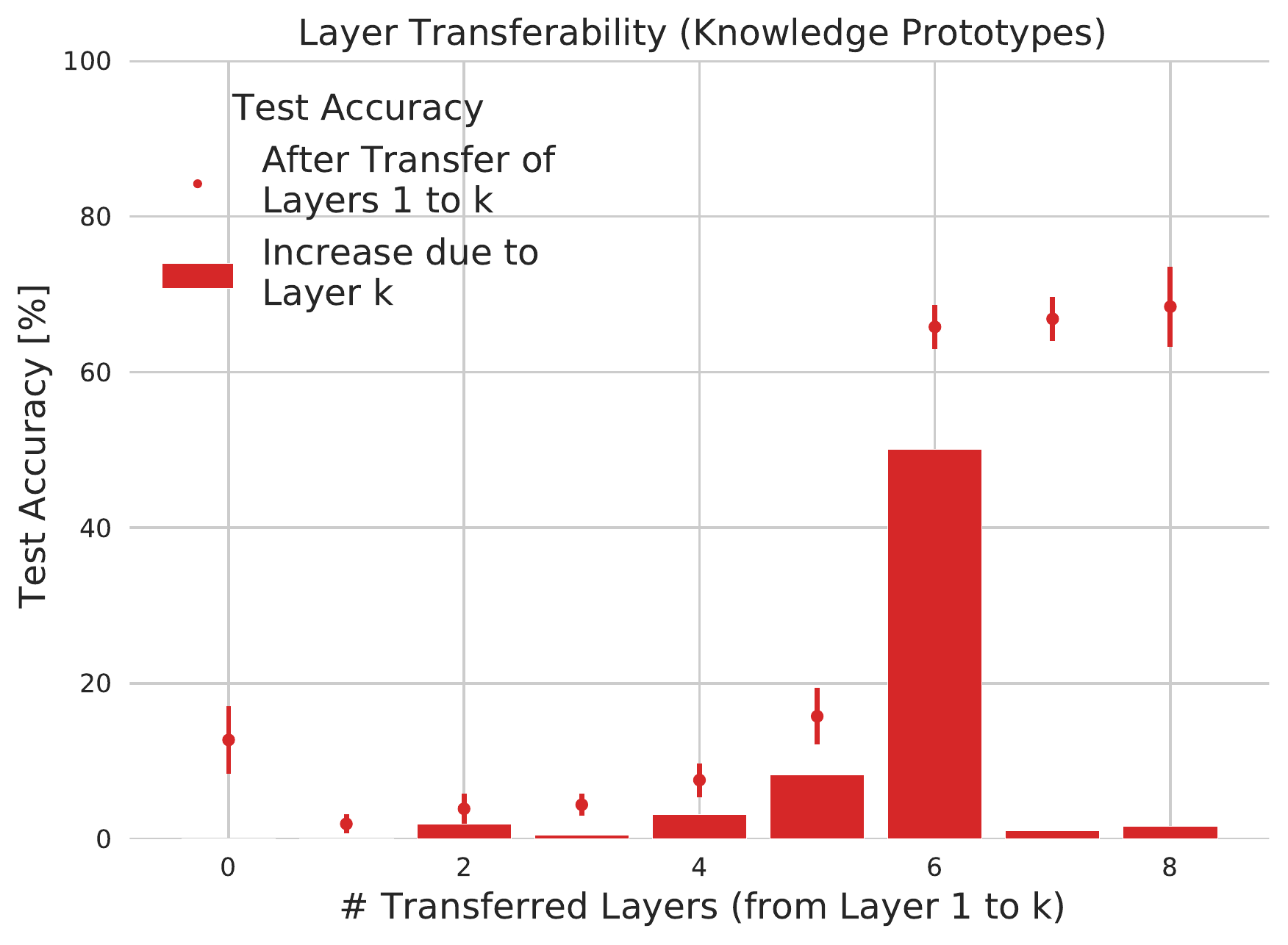}
    \caption{Knowledge-based Pre-Training}
    \label{fig:layer_transfer_prtr-know}
\end{subfigure}
\begin{subfigure}[t]{0.32\textwidth}
    \centering
    \includegraphics[width=\textwidth]{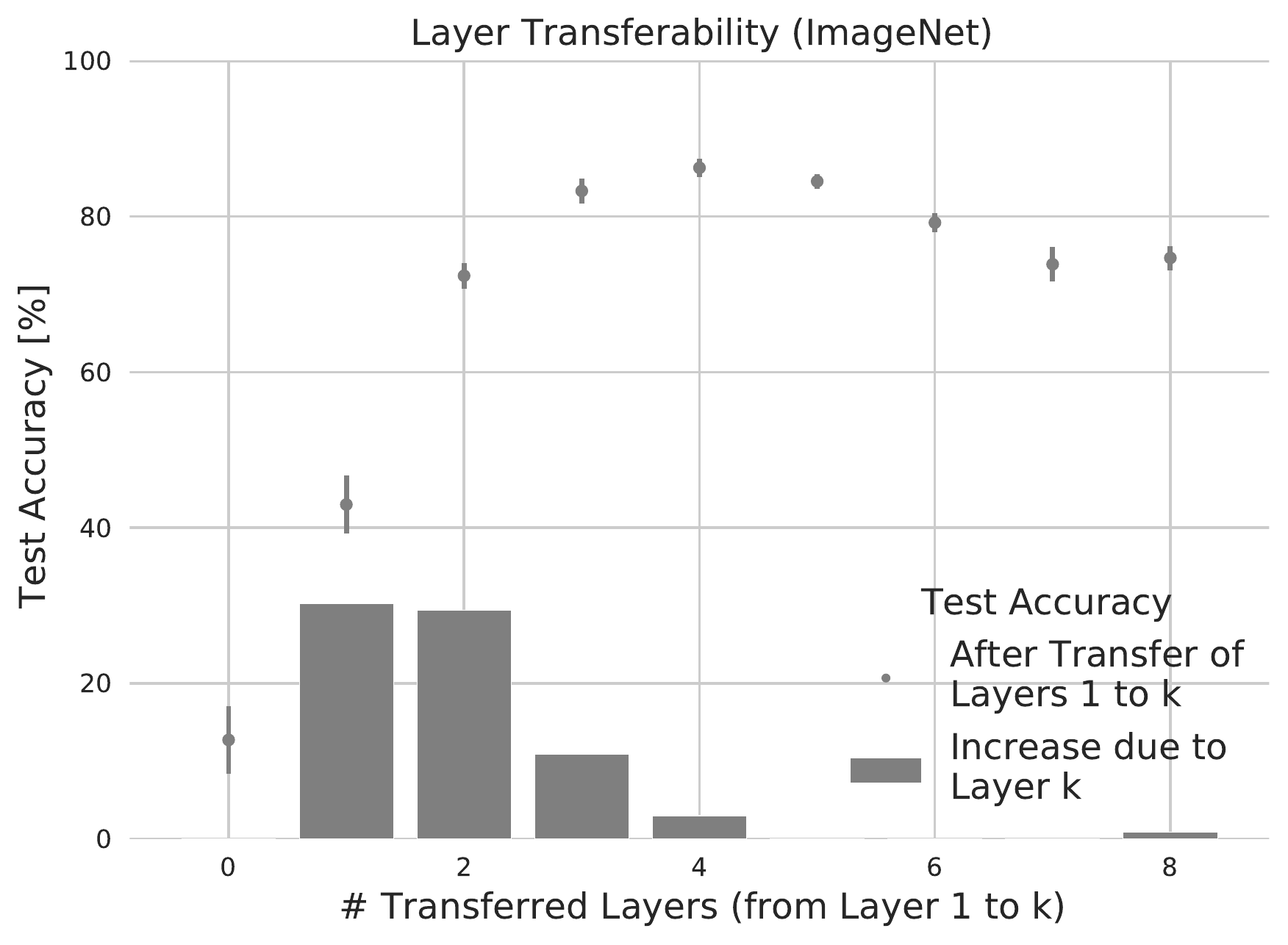}
    \caption{Data-based Pre-Training}
    \label{fig:layer_transfer_prtr-imn}
\end{subfigure}
\begin{subfigure}[t]{0.32\textwidth}
    \centering
    \includegraphics[width=\textwidth]{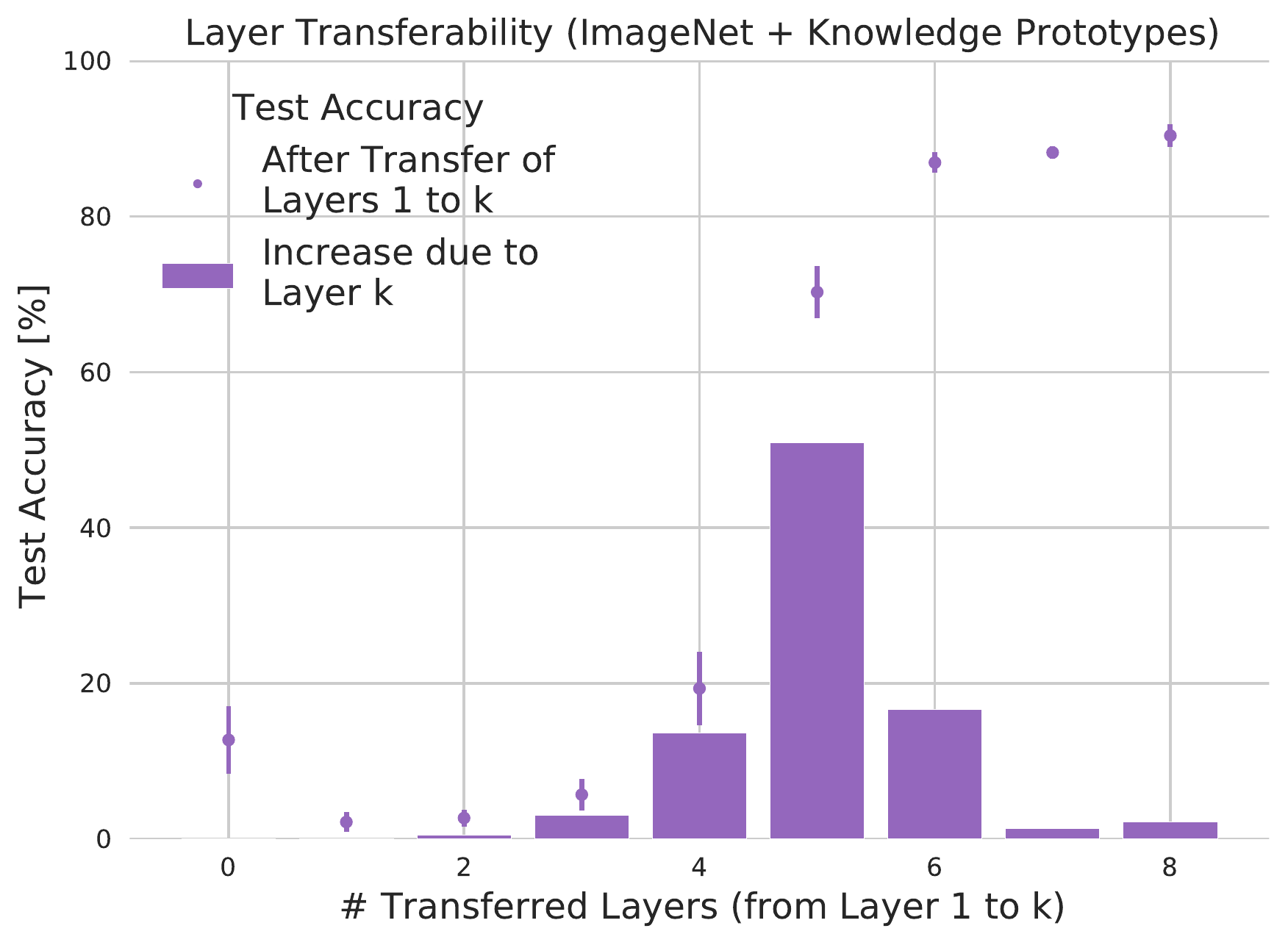}
    \caption{Data + Knowledge Pre-Training}
    \label{fig:layer_transfer_prtr-imn5-know}
\end{subfigure}
\caption{
Importance of individual network layers obtained by different pre-training types.
The figure shows results for training the AlexNet architecture, which has 8 layers. Here, after different pre-training schemes (\subref{fig:layer_transfer_prtr-know}) - (\subref{fig:layer_transfer_prtr-imn5-know}) only the first $k$ layers are transferred.
(\subref{fig:layer_transfer_prtr-know}) The network is pre-trained on knowledge prototypes using 100 augmented prototypes per class. 
(\subref{fig:layer_transfer_prtr-imn}) Results for a network that is pre-trained on the ImageNet dataset.
(\subref{fig:layer_transfer_prtr-imn5-know}) Combining both types by using a model pre-trained on ImageNet for the initialization of a subsequent pre-training on knowledge prototypes.
The experiment shows that knowledge-based pre-training has an additional and complementary effect to traditional data-based pre-training.
}
\label{fig:layer_transfer}
\end{figure*}
%%%%%%%%%%%%%%%%%%%%%%%%%%%%%%%%%%%%%%%%%%%%%%%%%%%%%%%%%%%%%%%%%%%%%%%%%%%%%%%%%%%%%%%%%%%%%%%%%%%%%%%%%%%%%%%%%%%%%%
%%%%%%%% Test on GTSRB - IMN Pr-Tr %%%%%%%%%%%%%%%%%%%%%%%%%%%%%%%%%%%%%%%%%%%%%%%%%%%%%%%%%%%%%%%%%%%%%%%%%%%%%%%%%%%%%%%%%%%%%%%
%%%%%%%%%%%%%%%%%%%%%%%%%%%%%%%%%%%%%%%%%%%%%%%%%%%%%%%%%%%%%%%%%%%%%%%%%%%%%%%%%%%%%%%%%%%%%%%%%%%%%%%%%%%%%%%%%%%%%%

\begin{table*}[t]%
\centering
\footnotesize
\begin{filecontents*}{tables/results_tex_table_gtsrb_gtsrb_alexnet-prtr_ep_10_acc_vs_trainset_size_LAYER_MODELS_no_highlighting.csv}
pre-training,74,592,4710,37890
(default init.),16.61 \pm 1.74,65.01 \pm 2.46,58.97 \pm 4.79,94.82 \pm 0.48
know. proto. (aug.),24.50 \pm 6.61,75.96 \pm 3.69,67.89 \pm 5.91,95.53 \pm 0.62
imagenet,28.50 \pm 1.12,76.31 \pm 1.71,62.25 \pm 3.75,96.92 \pm 0.34
imagenet + know. proto. (aug.),41.72 \pm 2.72,85.48 \pm 1.85,74.37 \pm 3.44,97.27 \pm 0.21
\end{filecontents*}
\pgfplotstabletypeset[
    col sep=comma,
    ignore chars={\$},% siunitx does not like $\pm$, but \pm is fine
    string type,
    % multicolumn header
    % every head row/.style={before row={\toprule & \multicolumn{1}{c}{} & \multicolumn{4}{c}{\tf{Test Accuracy [\%], for different Sizes of Training Data}}\\}, after row=\midrule,},
    % Define header 
    multicolumn names={l},% do not apply S format to header
    display columns/0/.style={column type = {l},column name={\tf{Pre-Training}}},
    display columns/1/.style={column type = {S},column name={\hspace{0.6em}\tf{$\approx$ 0.1\%}}},
    display columns/2/.style={column type = {S},column name={\hspace{0.6em}\tf{$\approx$ 1\%}}},
    display columns/3/.style={column type = {S},column name={\hspace{0.6em}\tf{$\approx$ 10\%}}},
    display columns/4/.style={column type = {S},column name={\hspace{0.6em}\tf{100\%}}},
    % rules
    every head row/.style={before row={\toprule}, after row={\midrule},},
    every last row/.style={after row=\bottomrule},
    % textbf for know. proto. labels
    every row no 1 column no 0/.style={postproc cell content/.style={@cell content=\tf{##1}}},
    every row no 3 column no 0/.style={postproc cell content/.style={@cell content=\tf{##1}}},
    % grey fonts
    every row no 0 column no 0/.style={postproc cell content/.style={@cell content=\gr{##1}}}
]{tables/results_tex_table_gtsrb_gtsrb_alexnet-prtr_ep_10_acc_vs_trainset_size_LAYER_MODELS_no_highlighting.csv}
\caption{
Test accuracies in \% for different train data sizes and pre-training on knowledge prototypes vs. pre-training on ImageNet.
Knowledge-based pre-training leads to additional improvements in test accuracy (up to 13\% for the small-data scenario).
}
\label{tab:results_imn_comparison}
\end{table*}

The goal of this experiment is to better understand the effect of informed pre-training.
For regular data-based pre-training, e.g., on ImageNet, the improvement results from the transfer of early neural network layers,  representing low-level, statistical features~(\cite{yosinski2014transferable}).
This raises the question, which network layers are responsible for performance gains when using knowledge prototypes.
We design an experiment to analyze the importance of individual network layers for transferring learning performance.
Our results reveal that for knowledge-based pre-training, the improvement arises from the transfer of mid to late network layers, which represent rather high-level, semantic features.

As before, we pre-train a neural network on knowledge prototypes but proceed training with only a specific part of the pre-trained model reusing it for informed initialization.
In particular, we only reuse the first $k$ layers.
I.e., the pre-trained model is truncated after layer~$k$.
The remaining layers are re-initialized with the default random procedure.
We use the GTSRB dataset and the AlexNet architecture, which has a total of eight layers resulting in nine partitioning cases.
Thus, $k=0$ corresponds to a random initialization of all layers and $k=8$ to deploying the full pre-trained model.
For each case, we perform a separate fine-tuning and then evaluate the test accuracy.
As the effect of pre-training becomes especially apparent in the early training phase, we deliberately evaluate the test accuracy after the first epoch.

Figure~\ref{fig:layer_transfer} shows the results of this transfer learning experiment for the small GTSRB subset (1\%) and for three different types of pre-training.
As the effect of pre-training becomes especially apparent in the early training phase, we deliberately evaluate the test accuracy after the first epoch for the full training set, which correspond to 64 Epochs for the small data set.
Each subfigure presents the test accuracy of the fine-tuned model if layers 1 to $k$ are initialized with the pre-trained model.
The inserted bar charts highlight the performance gain due to the transfer of the layers preceding layer $k$.
Figure \ref{fig:layer_transfer_prtr-know} shows the results for our pre-training on knowledge prototypes.
We learn that the test accuracy of the fine-tuned model is improved when the deep layers containing high-level semantic features are transferred.
In the literature, the effect of pre-training visual models is attributed to the transfer of early network layers while our results show that with informed initialization high-level features from the late layers are carried over.

Figure \ref{fig:layer_transfer_prtr-imn} shows the results for the same experimental procedure but for pre-training on ImageNet.
We see that the performance gain is the result of a transfer of the early network layers accounting for general low-level image features (as e.g. described by~\cite{yosinski2014transferable}).
Figure~\ref{fig:layer_transfer_prtr-imn} also reveals that the maximum test accuracy is achieved when the first five layers are transferred while transferring further layers even leads to a slight deterioration.

Our results suggest that the two pre-training types have complementary strengths.
Data-based pre-training is well suited for transferring low-level features whereas our knowledge-based pre-training lends itself well for informed transfer of high-level semantic features.
We therefore investigate a third pre-training method that consists of two pre-training phases initializing the first five layers with a regular ImageNet-pretraining and subsequently pre-training on knowledge prototypes.
The layer transferability for this approach is illustrated in Figure~\ref{fig:layer_transfer_prtr-imn5-know}.
As it depicts the contribution of the layers from knowledge-based pre-training, we again measure the largest gains truncating the late layers, albeit with a slight shift to the middle.
However, even more remarkable, a subsequent knowledge-based pre-training can further improve pre-training on ImageNet as reported in Table~\ref{tab:results_imn_comparison}.

\subsection{Informed Learning}
\label{sec:54_informed}
Finally, we have also tested other forms of informed learning. We injected the knowledge prototypes in the training data itself and trained models concurrently on both. We also combined informed pre-training and this informed learning. The results are illustrated in Figure~\ref{fig:mixed}.
The complete results can be found in the Appendix.
It shows that not only pre-training, but also the concurrent learning can improve test accuracy. However, for the training on very small MNIST data, the pure pre-training leads to the better results. Otherwise, combined informed pre-training and learning yields the best results.
All in all, this experiment shows that the proposed utilization of knowledge prototypes leads to significant improvements and can go way beyond pre-training.
An advantage of pre-training, which is not reflected in this experiment, is the following: Knowledge can also change over time. If we first pre-train on knowledge and then refine to real data observations, the inductive bias is less strong and still allows the flexibility to adapt.

\begin{figure*}[t]%[!htb]
\centering
\begin{subfigure}[b]{0.32\textwidth}
    \includegraphics[width=\textwidth]{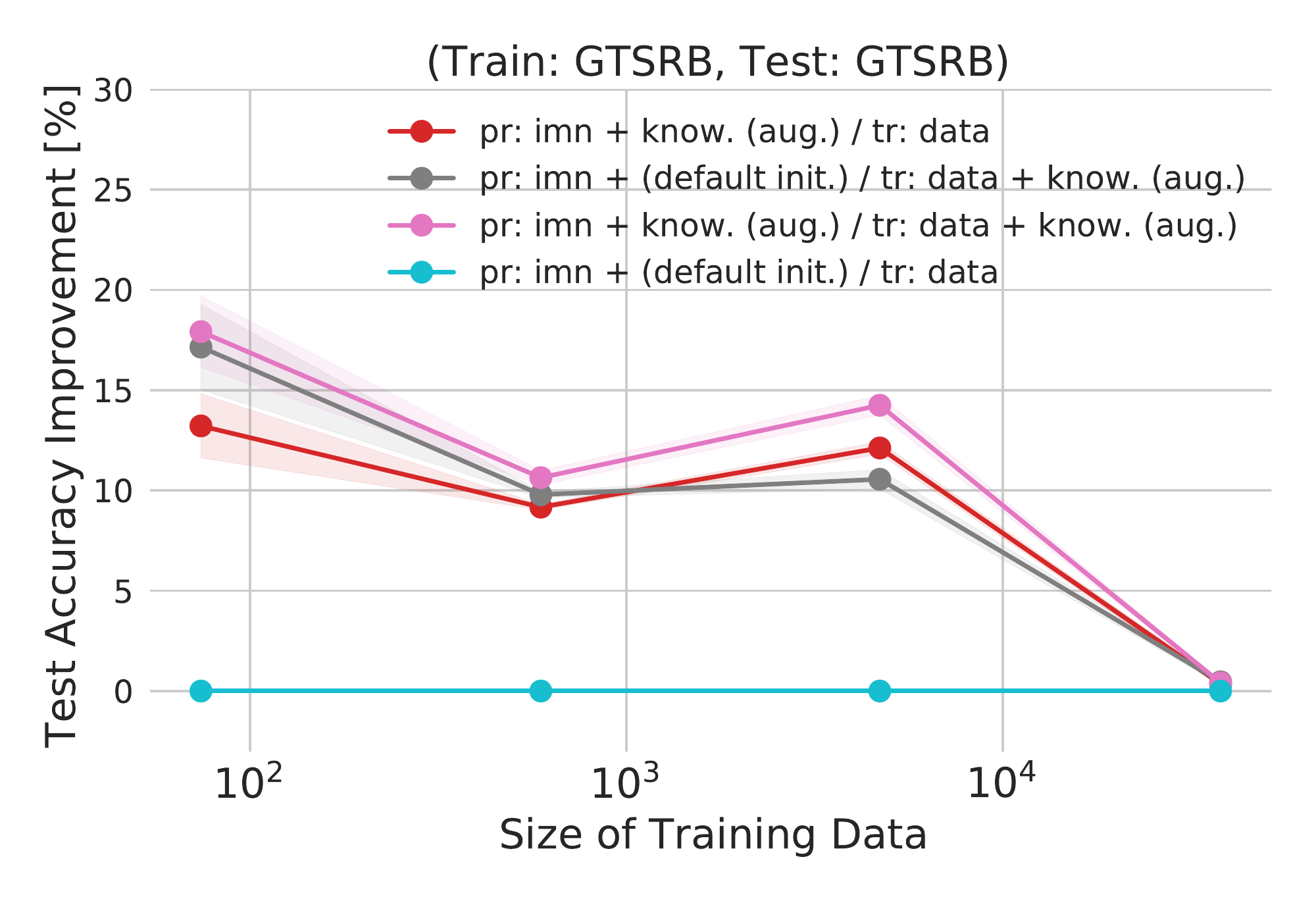}
    %\caption{GTSRB}
    %\label{fig:gtsrb_mixed}
\end{subfigure}
\begin{subfigure}[b]{0.32\textwidth}
    \includegraphics[width=\textwidth]{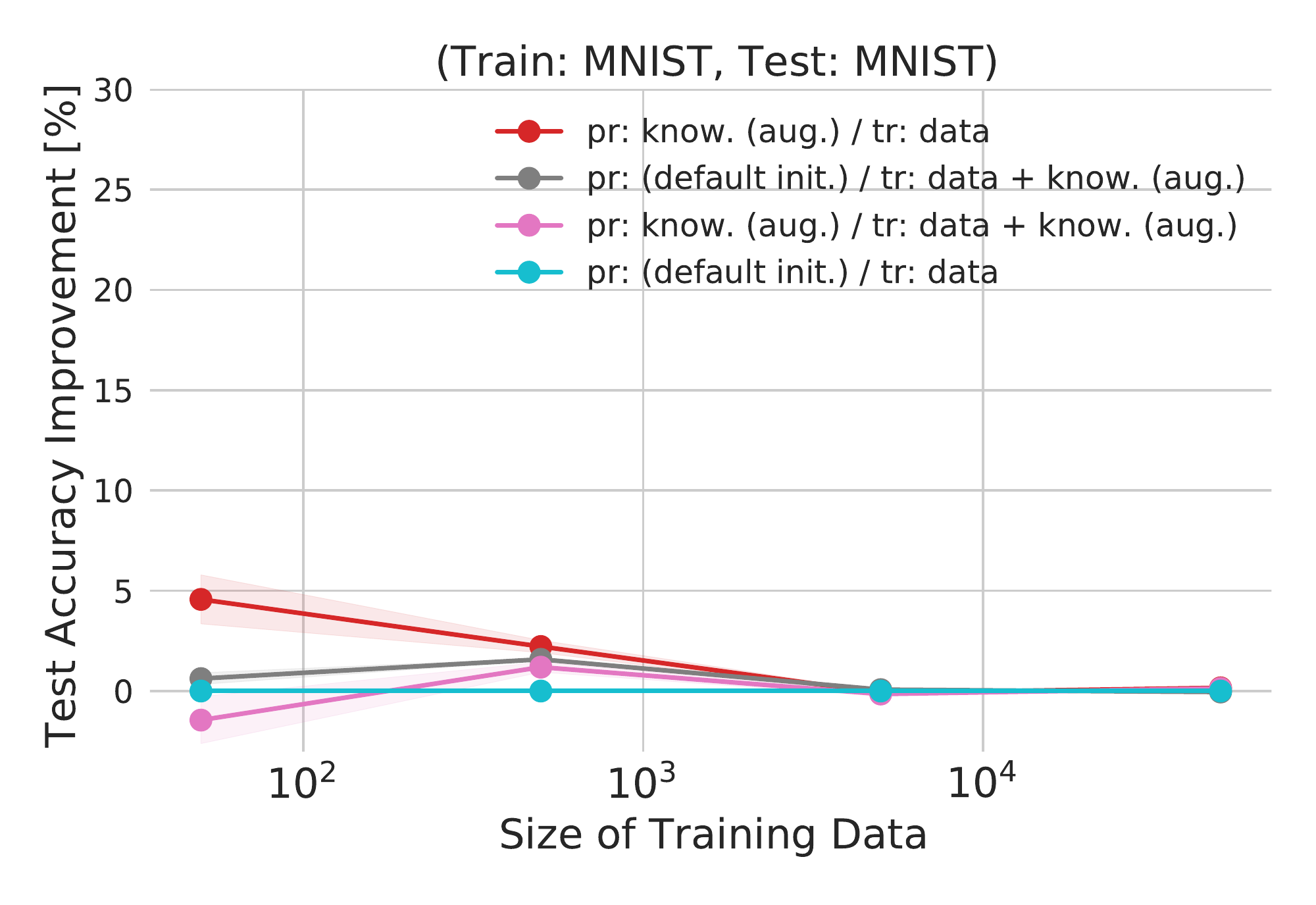}
    %\caption{ MNIST}
    %\label{fig:mnist_mixed}
\end{subfigure}
\begin{subfigure}[b]{0.32\textwidth}
    \includegraphics[width=\textwidth]{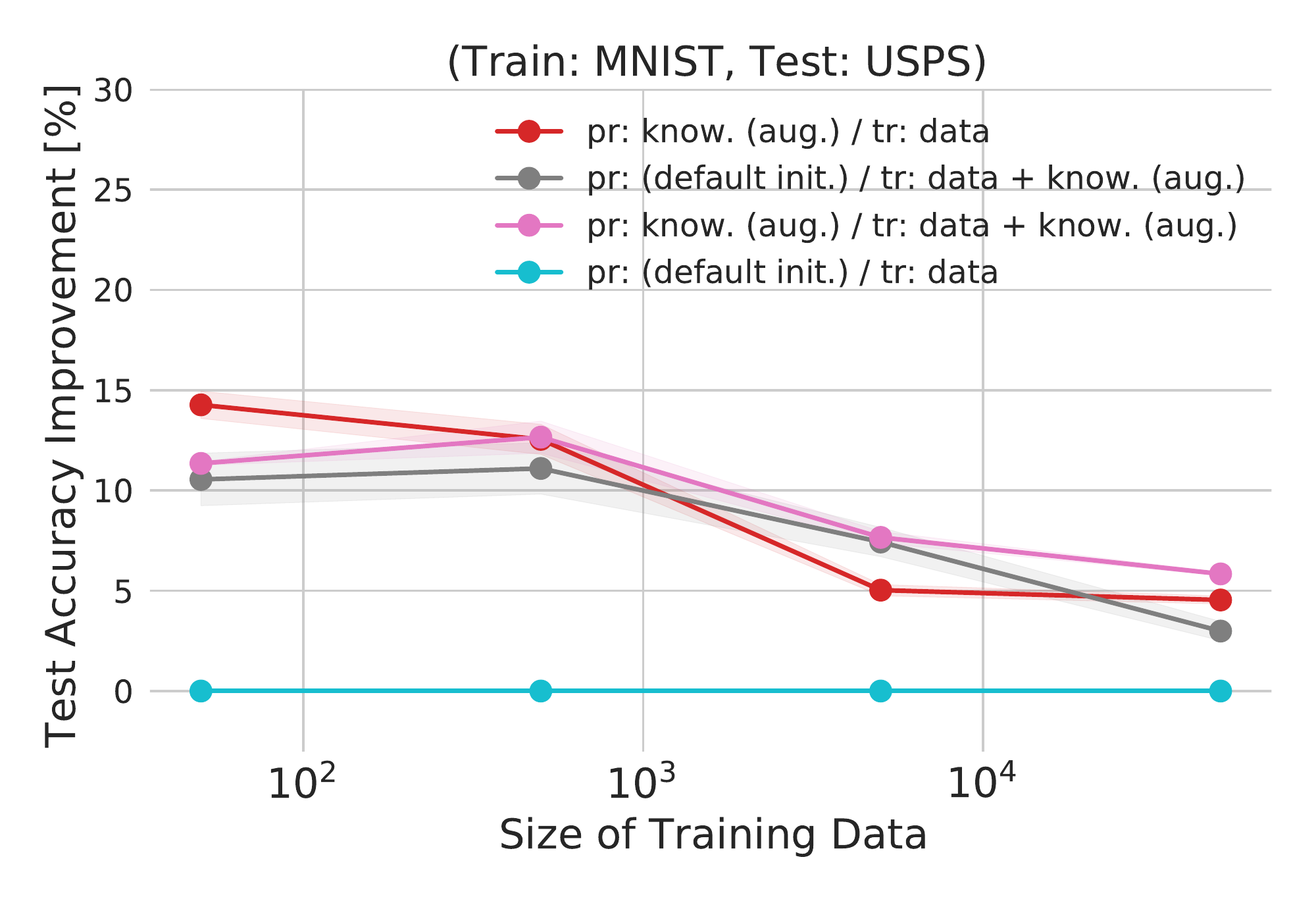}
    %\caption{USPS}
    %\label{fig:usps_mixed}
\end{subfigure}
\caption{
Improvements in test accuracy for informed pre-training (pre-training on prototypes), informed learning (concurrent learning on data and prototypes), as well as combined informed pre-training and learning.
}
\label{fig:mixed}
\end{figure*}

\section{Conclusion}
\label{sec:conclusions}
We proposed informed pre-training on prior knowledge and found that it speeds up the training process and improves both out-of-distribution robustness and generalization, in particular in situations with little training data.
We showed that knowledge-based pre-training has novel and complementary strengths to traditional data-based pre-training. While for the latter the transfer learning mainly affects the early network layers, for informed transfer learning the late network layers, which often represent semantic features, are most relevant.
This is an interesting insight, which also seems to agree with the notion from cognitive sciences~(\cite{minsky1991logical, chipman2012foundations}) that semantic prior knowledge is most useful when the lower-level connections have already been learned.
A huge advantage of our method is that not domain specific and can be applied to various knowledge representations types and domains.
However, a potential challenge is the transformation of formal knowledge into prototypes. For scientific and visual domains technique like rendering and simulation are available. For other domains like language modelling the transformation might require more research.
Therefore as future research we see the application to more sophisticated knowledge types.
For example, one could use graph structures or high-dimensional geometrical models that are otherwise used for synthetic data generation (such as by \citet{schwarz2020stillleben}) and employ them as knowledge prototypes for informed initialization.

%%%%%%%%%%%%%%%%%%%%%%%%%%%%%%%%%%%%%%%%%%%%%%%%%%%%%%%%%%%%%%%%%%%%%%%%%%%%%%%
%%%%%%%%%%%%%%%%%%%%%%%%%%%%%%%%%%%%%%%%%%%%%%%%%%%%%%%%%%%%%%%%%%%%%%%%%%%%%%%
% Acknowledgements
%%%%%%%%%%%%%%%%%%%%%%%%%%%%%%%%%%%%%%%%%%%%%%%%%%%%%%%%%%%%%%%%%%%%%%%%%%%%%%%
%%%%%%%%%%%%%%%%%%%%%%%%%%%%%%%%%%%%%%%%%%%%%%%%%%%%%%%%%%%%%%%%%%%%%%%%%%%%%%%

% \begin{ack}
% Use unnumbered first level headings for the acknowledgments. All acknowledgments
% More information about this disclosure can be found at: \url{https://neurips.cc/Conferences/2022/PaperInformation/FundingDisclosure}.

% Do {\bf not} include this section in the anonymized submission, only in the final paper. You can use the \texttt{ack} environment provided in the style file to automatically hide this section in the anonymized submission.
% \end{ack}

%%%%%%%%%%%%%%%%%%%%%%%%%%%%%%%%%%%%%%%%%%%%%%%%%%%%%%%%%%%%%%%%%%%%%%%%%%%%%%%
%%%%%%%%%%%%%%%%%%%%%%%%%%%%%%%%%%%%%%%%%%%%%%%%%%%%%%%%%%%%%%%%%%%%%%%%%%%%%%%
% References
%%%%%%%%%%%%%%%%%%%%%%%%%%%%%%%%%%%%%%%%%%%%%%%%%%%%%%%%%%%%%%%%%%%%%%%%%%%%%%%
%%%%%%%%%%%%%%%%%%%%%%%%%%%%%%%%%%%%%%%%%%%%%%%%%%%%%%%%%%%%%%%%%%%%%%%%%%%%%%%

\newpage
\bibliography{references}
\bibliographystyle{abbrvnat}

%%%%%%%%%%%%%%%%%%%%%%%%%%%%%%%%%%%%%%%%%%%%%%%%%%%%%%%%%%%%
%%%%%%%%%%%%%%%%%%%%%%%%%%%%%%%%%%%%%%%%%%%%%%%%%%%%%%%%%%%%
% Appendix
%%%%%%%%%%%%%%%%%%%%%%%%%%%%%%%%%%%%%%%%%%%%%%%%%%%%%%%%%%%%
%%%%%%%%%%%%%%%%%%%%%%%%%%%%%%%%%%%%%%%%%%%%%%%%%%%%%%%%%%%%
\todo{Add appendix}

% \newpage
% \appendix

% \section{Appendix}
% \input{7_appendix}

% Optionally include extra information (complete proofs, additional experiments and plots) in the appendix.
% This section will often be part of the supplemental material.

\end{document}